\documentclass[10pt,twocolumn,letterpaper,table]{article}

\usepackage{iccv}              

\definecolor{iccvblue}{rgb}{0.21,0.49,0.74}
\usepackage[pagebackref,breaklinks,colorlinks,allcolors=iccvblue]{hyperref}

\usepackage{url}
\usepackage{adjustbox}
\usepackage{epsfig}
\usepackage{graphicx}
\usepackage{amsmath}
\usepackage{amssymb}
\usepackage{wrapfig} 

\usepackage{xcolor}
\usepackage{subfloat}
\usepackage{multirow}
\usepackage{booktabs}
\usepackage{makecell}
\usepackage{mdframed}
\usepackage{colortbl}
\usepackage[accsupp]{axessibility}

\definecolor{darkgreen}{HTML}{228B22}
\definecolor{darkgray}{HTML}{757575}

\newcolumntype{P}[1]{>{\centering\arraybackslash}p{#1}}

\usepackage[normalem]{ulem}

\def\Method{\textbf{Oasis}}

\newcommand{\authorskip}{\hspace{2.5mm}}
\newcommand{\affilationskip}{\hspace{5mm}}
\usepackage{tocloft}

\makeatletter
\renewcommand{\numberline}[1]{%
  \@cftbsnum #1\@cftasnum\hspace*{1em}\@cftasnumb%
}
\makeatother

\def\paperID{6406} 
\def\confName{ICCV}
\def\confYear{2025}

\usepackage{listings}

\title{\Method: One Image is All You Need for Multimodal Instruction Data Synthesis}

\author{
\hspace{-2mm}Letian Zhang\thanks{Work done during internship at Bytedance.}{\normalsize\textsuperscript{$*1,2$}} \authorskip
Quan Cui{\normalsize\textsuperscript{$2$}} \authorskip
Bingchen Zhao{\normalsize\textsuperscript{$3$}} \authorskip
Cheng Yang{\normalsize\textsuperscript{$2$}} \vspace{2mm} \\
\normalsize
\textsuperscript{$1$}Tongji University \affilationskip
\textsuperscript{$2$}Bytedance \affilationskip
\textsuperscript{$3$}University of Edinburgh
}

\begin{document}
\maketitle

\begin{abstract}

  The success of multi-modal large language models (MLLMs) has been largely attributed to the large-scale training data. However, the training data of many MLLMs is unavailable due to privacy concerns. The expensive and labor-intensive process of collecting multi-modal data further exacerbates the problem. Is it possible to synthesize multi-modal training data automatically without compromising diversity and quality? In this paper, we propose a new method, \Method, to synthesize high-quality multi-modal data with only images. \Method~breaks through traditional methods by prompting only images to the MLLMs, thus extending the data diversity by a large margin.
  Our method features a delicate quality control method which ensures the data quality.
  We collected over 500k data and conducted incremental experiments on LLaVA-NeXT. 
  Extensive experiments demonstrate that our method can significantly improve the performance of MLLMs. The image-based synthesis also allows us to focus on the specific-domain ability of MLLMs. 
  Code and dataset are publicly available at  \url{https://github.com/Letian2003/MM_INF}.

\end{abstract}

\vspace{-1em}

\section{Introduction}
\label{sec:intro}
Multi-modal large language models (MLLMs) have become a popular research topic in recent communities due to their superior performance in various multi-modal tasks. The success of MLLMs relies heavily on the large-scale training data, which directly compose the model's knowledge base. However, the lack of multi-modal training data has been a bottleneck for the development of MLLMs, since the training data of top MLLMs are typically private. Therefore, an effective way to synthesize high-quality multi-modal data has been a long-standing challenge for the community.

Previous studies have presented some effective methods to synthesize multi-modal data with low cost. LLaVA~\cite{llava} takes a GPT-assisted method to generate multi-modal instruction-following data based on existing image-pair data. 
ALLAVA~\cite{allava} achieves data synthesis using a captioning-then-QA fashion with the assistance of GPT-4V.
Some recent research focuses on the diversity and complexity of data. For example, MMEvol~\cite{mmevol} is an image-text instruction evolution framework, which iteratively enhances instruction diversity in multiple designed domains. 

  \begin{figure}[t]
    \centering
    \includegraphics[width=1\linewidth]{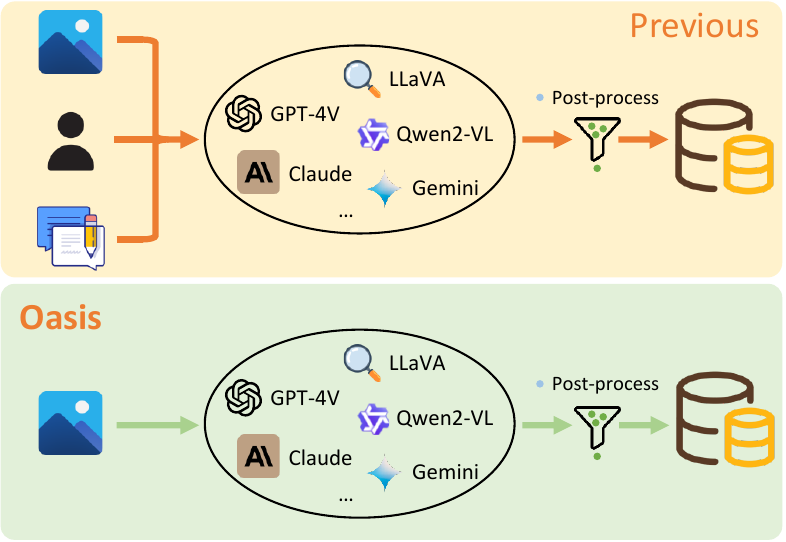}
    \vspace{-1.5em}
    \caption{\textbf{Comparison of previous methods and our proposed \Method~framework for multi-modal data synthesis.} 
    Previous approaches rely on an input image, complex prompts for generating text, and human labor. Responses are generated from advanced LLMs and MLLMs~(\eg, GPT-4V~\cite{gpt4} and Qwen2-VL~\cite{qwen_vl,qwen2_vl}). 
    Interestingly, our proposed \Method~requires only a single image to generate multi-modal instruction-following data, showing great simplicity and practical value. }
    \label{fig:intro}
    \vspace{-1.2em}
    
  \end{figure}

We analyze the existing synthesis approach and concluded three main concerns across this line of work: \textbf{(1) Constant pipeline compromises data diversity.} Invariable prompts and fixed flow constrains the data scope and homogenizes the difficulty, leading to an incomprehensive model. \textbf{(2) Insufficient control on data quality.} It is hard to synthesize high-quality data which could significantly improve the representation ability of MLLMs. Most studies adopt a caption-based GPT generation strategy to mitigate this problem. \textbf{(3) Complicated framework involves human engagement.} A self-contained synthesis framework typically requires human efforts to design data patterns or prompts, which makes data synthesis troublesome.

In response to these concerns, we propose \Method~(\underline{\textbf{O}}ne image is \underline{\textbf{a}}ll you need for multimodal data synthe\underline{\textbf{sis}}), a novel and straightforward method to create high-quality and diverse multi-modal data with only images. Inspired by Magpie~\cite{magpie}, we break the traditional input tokens and entice a strong MLLM to generate self-aligned instructions. No single text prompt is required and the only input to the MLLM is the image, which could be easily obtained from the web. The auto-regressive nature of MLLMs leads them to generate diverse instructions based on their own knowledge base. We dive into the property of good instructions and carefully design several standards to filter out low-quality data.

\Method~is a simple yet effective method for multi-modal data synthesis that takes only visual content as prior knowledge. This inherent image-based nature makes the generated data domain heavily dependent on the image domain. We take advantage of this characteristic and produce domain-oriented multi-modal data by controlling the source of images, while not compromising the data quality and diversity. 
To validate the effectiveness of our method, we collect over 500k~\Method~data and conduct extensive experiments on LLaVA-NeXT~\cite{llava_next}. 
The results demonstrate that incorporating our synthesized data into the original training set significantly enhances MLLM performance across 14 benchmarks with different backbones.
Moreover, our method outperforms existing synthesis approaches by a large margin.
We conduct a series of ablation studies to further verify the effectiveness of our quality control strategy. 
Additionally, we perform a case study in the OCR domain to showcase our method's capability in domain-specific tasks.

We summarize our contributions as follows:
\begin{enumerate}
    \item A novel and straightforward method~\Method~is proposed to synthesize multi-modal data of high diversity and quality, which only requires images as input. 
    \item By concluding the property of high-quality multi-modal data, we handcraft an array of quality control techniques to ensure the data quality.
    \item Extensive experiments demonstrate that~\Method~data effectively enhance MLLM capabilities across different backbones and outperform other synthesis methods.
    \item Over 500k~\Method~data and consequent models will be made publicly available, hoping to facilitate future research in this field.
\end{enumerate}


  \begin{figure*}[t]
    \centering
    \includegraphics[width=1\linewidth]{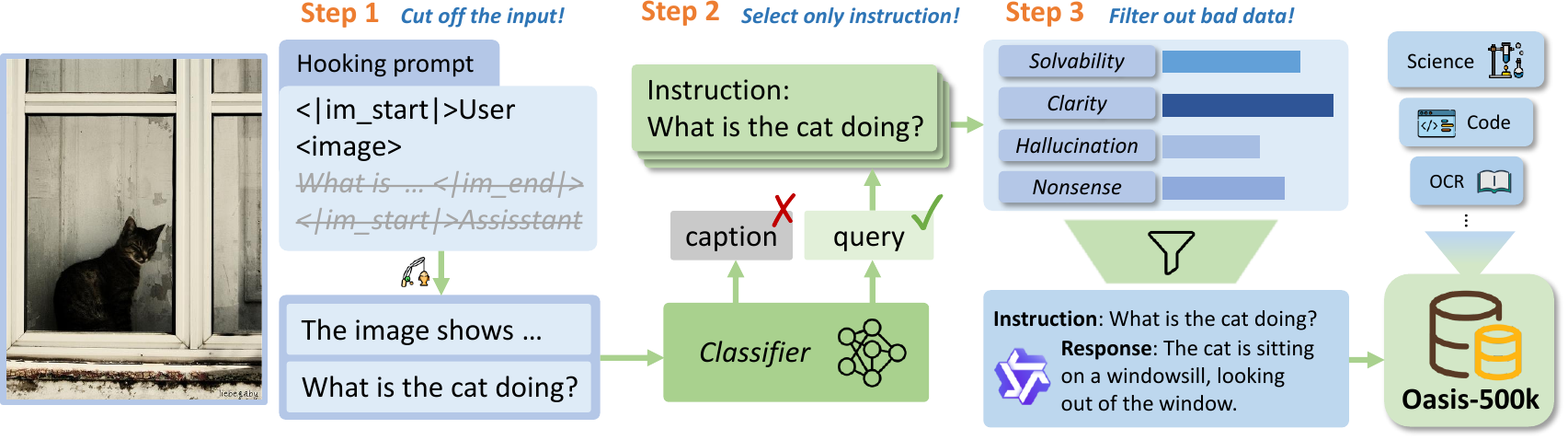}
    \vspace{-1.5em}
    \caption{\textbf{Detailed \Method~pipeline.} This figure illustrates the full process of data synthesis with \Method. The pipeline consists of three steps: data synthesis, data categorization, and instruction quality control. In Step 1, we break the traditional input tokens and entice a strong MLLM to generate instructions based on the image. In Step 2, we filter out non-instruction-following data by an LLM. In Step 3, a quality control mechanism is proposed to ensure the remained instructions are reasonable and high-quality. \Method~exhibits a straightforward and efficient way to synthesize multi-modal training data with low demands~(\ie, a single image input). The empirical results show that our method can significantly improve the performance of MLLMs.}
    \vspace{-.5em}
    \label{fig:method}
  \end{figure*}

\section{Related Work}
\label{sec:Related_Work}

\subsection{Multi-modal Large Language Models}
Multi-modal large language models (MLLMs) have achieved remarkable success in various vision-language tasks, \eg, GPT-4o~\cite{gpt4}, LLaVA series~\cite{llava,llava1_5,llava_next,llava_ov}, InternVL series~\cite{internvl,internvl1_5,internvl2_5} and Qwen-VL series~\cite{qwen_vl,qwen2,qwen2_vl,qwen2_5}. Popular MLLM architecture is composed of three components: a vision encoder, a large language model, and a projector.
Contrastive Language-Image Pre-training~(CLIP)~\cite{clip} vision encoders are widely used and incorporated with diverse LLMs~\cite{qwen2,qwen2_5,llama3,gpt3,gpt4}. The projector can be a compact module based on cross-attention layers or MLPs.
The multi-modal pre-training paradigm has been widely studied in recent years. A common multi-modal training process~\cite{cambrian} is typically divided into two stages: pre-training and fine-tuning. 
In the pre-training phase, the model is trained on a large-scale multi-modal caption dataset, and the objective is to align representations of different modalities.
The pre-training dataset is usually collected from the web and mainly consists of captions. 
In the fine-tuning stage, the model is fine-tuned on a specific downstream task, which involves a large amount of instruction-following data. 
The multi-modal pre-training paradigm has been proven to be effective in improving MLLM performance on various vision-language tasks, \eg, general question answering, OCR and visual reasoning.

\subsection{Multi-modal Data Synthesis}
The lack of multi-modal training data has been a bottleneck for the development of MLLMs. To address this issue, researchers have proposed various methods to synthesize multi-modal data in both LLM and MLLM communities~\cite{magpie,mminstruct,mme,alpaca,vicuna2023,selfinstruct,zhu2023minigpt,chen2024sharegpt4v,liu2023mitigating}. 

Multi-modal synthesis methods can be roughly divided into two categories: rule-based methods and generation-based methods. Rule-based methods synthesize multi-modal data based on predefined rules, while generation-based methods generate multi-modal data using generative models. Rule-based methods are usually simple and efficient, but they may lack diversity. 
Data synthesis methods in the multi-modal domain often take inspiration from the works focusing on generating language-only instruction data, example works include~\cite{xu2024wizardlm,alpaca,vicuna2023,selfinstruct}, leveraging the capability of the GPT4 model, Alpaca and Vicuna models~\cite{alpaca,vicuna2023} leverage the self instruct method to generate a large number of instruction data to improve the instruction following ability of base models.
Evol-Instruct~\cite{xu2024wizardlm} takes a step further by using evolution-based prompts to improve the diversity of the generated instruction data.
ALLaVA~\cite{allava} provides high-quality multi-modal data by handcraft efforts to rewrite and polish instruction-following data. Cambrian-1~\cite{cambrian} revises classical computer vision datasets by designing specific text as instructions.
Generation-based methods can generate diverse multi-modal data, but they may require a large amount of computational resources. For example, LLaVA's authors create LLaVA-Instruct~\cite{llava} by using advanced LLMs to label data. MMInstruct~\cite{mminstruct} employs state-of-the-art MLLMs to generate data. MMEvol~\cite{mmevol} and VILA2~\cite{vila2} study the prompt evolution technique to create complex and diverse multi-modal training data.
The majority of existing methods involve complex prompt designs and human labor in the synthesis procedure. 
The most related work to ours is a prompt-free text data synthesis work Magpie~\cite{magpie}, and further discussion of relations between Magpie and our method is provided in Appendix A.1. 
Inspired by this study, we propose a simple yet effective method named~\Method, which generates multi-modal instruction-following data with a single image.

\section{Methodology}

\subsection{Motivation}
  Automated multi-modal data generation often suffers from inadequate data quality and diversity, while demanding heavy human involvement. Previous studies have presented impressive methods to eliminate these obstacles. 
  However, we believe that the synthesis process can be performed in a more efficient but straightforward way, without compromising data quality and diversity. 
  In the following, we present a novel method named~\Method, which synthesizes multi-modal instruction-following data with only one image.

\subsection{\Method:~A Novel Synthesis Method}
\label{sec:synthesis}
Our \Method~data synthesis pipeline consists of three steps, and the detailed flow is presented in~\cref{fig:method}. In the first step, we extract the response of a fully optimized MLLM with a ``hooking'' prompt. In the second step, we categorize the response of MLLM with the assistance of an LLM. In the third step, we design a quality control mechanism to filter inferior responses with an LLM. In the following, we present the details of the above steps.

  \paragraph{Step 1: Data synthesis with ``hooking'' prompt.}
  We use an MLLM to extract the response to the image input. The typical input to an MLLM can be broken into four components: the pre-query template, the visual content, the instruction, and the post-query template. In the case of Qwen2-VL, an input can be ``\texttt{\textcolor{blue!80}{<|im\_start|>User <image>}\textcolor{red!80}{Describe the image.<|im\_end|> <|im\_start|>Assisstant}}". 
  Among these components, the pre-query template and the post-query template are fixed, dividing the boundary of the user and the assistant part. Therefore, the response of the MLLM can be formulated as $Resp = \Theta(vision, instruction)$, where $\Theta$ represents the MLLM. In \Method, the \textcolor{blue!80}{blue} part is considered \textbf{a hook} for the \textcolor{red!80}{red} part. As shown in~\cref{fig:method}, we intentionally remove the query part and the post-query template, making the MLLM generate the instruction autoregressively based only on the hooking prompt, \ie, $Inst = \Theta(vision)$. By this means, the generated instructions are free from manually designed prompts, thus more diverse. 
  
  \paragraph{Step 2: Data categorization.}
  \label{para:categorization}
  During synthesis, we observe that the generated data can be roughly divided into two categories: instruction-following and caption. The data can be either instruction-following, where the instruction guides the user to perform a specific task, or caption, where the instruction describes the content of the image. This phenomenon can be explained by the interleaved MLLM image-text training process. To select only instruction data, we design a categorization mechanism to classify the data into two categories as depicted in~\cref{fig:method}. Specifically, we prompt an LLM as a classifier to predict the category of the data. If the data contains instructions, it is classified as instruction-following data and an instruction is extracted. Otherwise, it is classified as caption data and deserted. We use few-shot to improve the classification accuracy and the full prompts can be found in Appendix B.1. 
  
  \paragraph{Step 3: Instruction quality control.}
  \label{para:quality}
  With in-depth observation of previous works, we believe that the quality of the instruction is the key to the success of the synthesized data. Therefore, after a thorough analysis of the properties of high-quality instructions, we summarize four important characteristics. \textbf{(1) Solvability.} Does the image provide all the necessary information to answer the question comprehensively. \textbf{(2) Clarity.} How precisely the question conveys its intent and whether it allows for a definitive answer. \textbf{(3) Hallucination.} Alignment between the question's content and the actual content of the image. \textbf{(4) Nonsense.} Whether the question is grammatically correct, coherent, and semantically meaningful. We rate all instructions according to these four dimensions on a scale of 1 to 5 and filter out low-quality instructions. The first 3 standards are evaluated by an MLLM, while the last one is evaluated by an LLM, since LLM is more sensitive to the language quality. Specifically, we handcrafted elaborate scoring criteria for each dimension, and the prompts are listed in Appendix B.2. Evaluation cases are provided in~\cref{fig:inst_qc_case}. Response quality control is proven instead to be ineffective in~\cref{para:resp_qc}.

\paragraph{\Method-500k data. }
We select Cambrian-10M~\cite{cambrian} dataset as the image source for reproduction, and collect 500k data, named \Method-500k. Due to the great scalability, \Method~data could be easily scaled as long as there are enough images. Therefore, the size of the data can grow linearly over time.

\subsection{\Method-500k Data Attributes}
\label{attributes}
In this part, we provide in-depth explorations of the attributes of synthetic data for better understanding \Method.

  \paragraph{\Method~data has long text lengths of instructions and responses.} In \cref{tab:length}, we report the average length and standard deviation of the instruction and response data in LLaVA-NeXT and \Method-500k. The results show that the average length of \Method~data is about double that of LLaVA-NeXT, indicating that our data is more informative and detailed. More details of the input image or more complicated reasoning contents can be provided by \Method~data. The standard deviation of \Method~data is also higher than that of LLaVA-NeXT, suggesting that our data is more diverse and takes variable forms.

  \begin{table}[t]
    \caption{\textbf{Data length statics.} This table presents a comparison of data length statistics between LLaVA-NeXT and \Method-500k. Both instructions and responses in our data are longer and more variable than those in LLaVA-NeXT. This indicates that our data is more informative and diverse.}
    \centering
    \small
    \begin{tabular}{lcccc}
    \toprule
    \multirow{2}{*}{Data source} & \multicolumn{2}{c}{Ave. Length} & \multicolumn{2}{c}{Std. Deviation} \\
     \cmidrule(r){2-3} \cmidrule(r){4-5}
     & Inst. & Resp. & Inst. & Resp. \\
    \midrule
    LLaVA-NeXT & 45.24 & 34.16 & 55.03 & 185.30 \\
    \Method & 76.80 & 71.16 & 375.76 & 529.34 \\
    \bottomrule
    \end{tabular}
    \label{tab:length}
\end{table}

  \begin{figure}[t]
    \centering
    \includegraphics[width=1\linewidth]{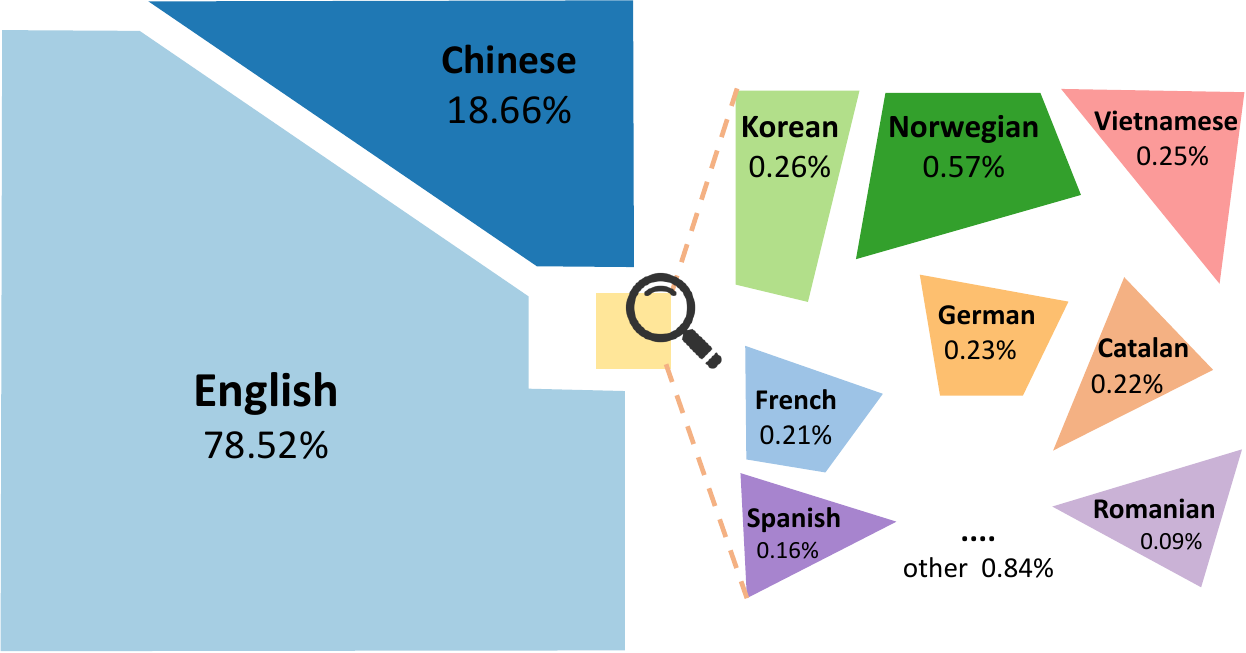}
    \caption{\textbf{Language type breakdown.} The distribution of language types in \Method~data. English takes up the majority, while other languages are also well-represented. In total, 46 language types are included in the dataset.}
    \label{fig:language}
    
  \end{figure}

      \begin{figure}[t]
    \centering
    \subfloat[LLaVA-NeXT data]{
      \begin{minipage}[b]{0.22\textwidth}
        \includegraphics[width=1\textwidth]{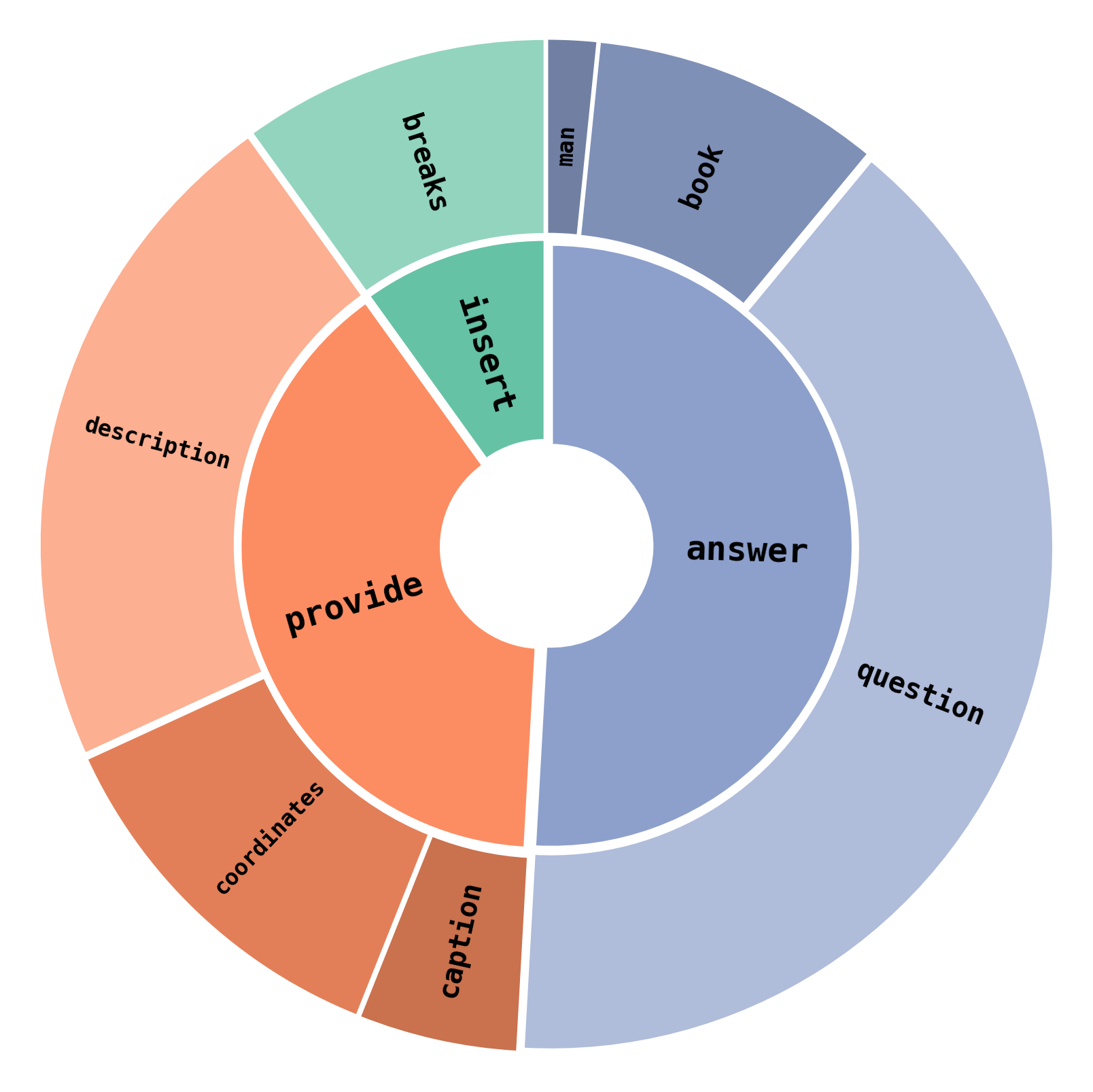}
      \end{minipage}
      \label{fig:direct}
      }
      \subfloat[\Method~data]{
      \begin{minipage}[b]{0.22\textwidth}
        \includegraphics[width=1\textwidth]{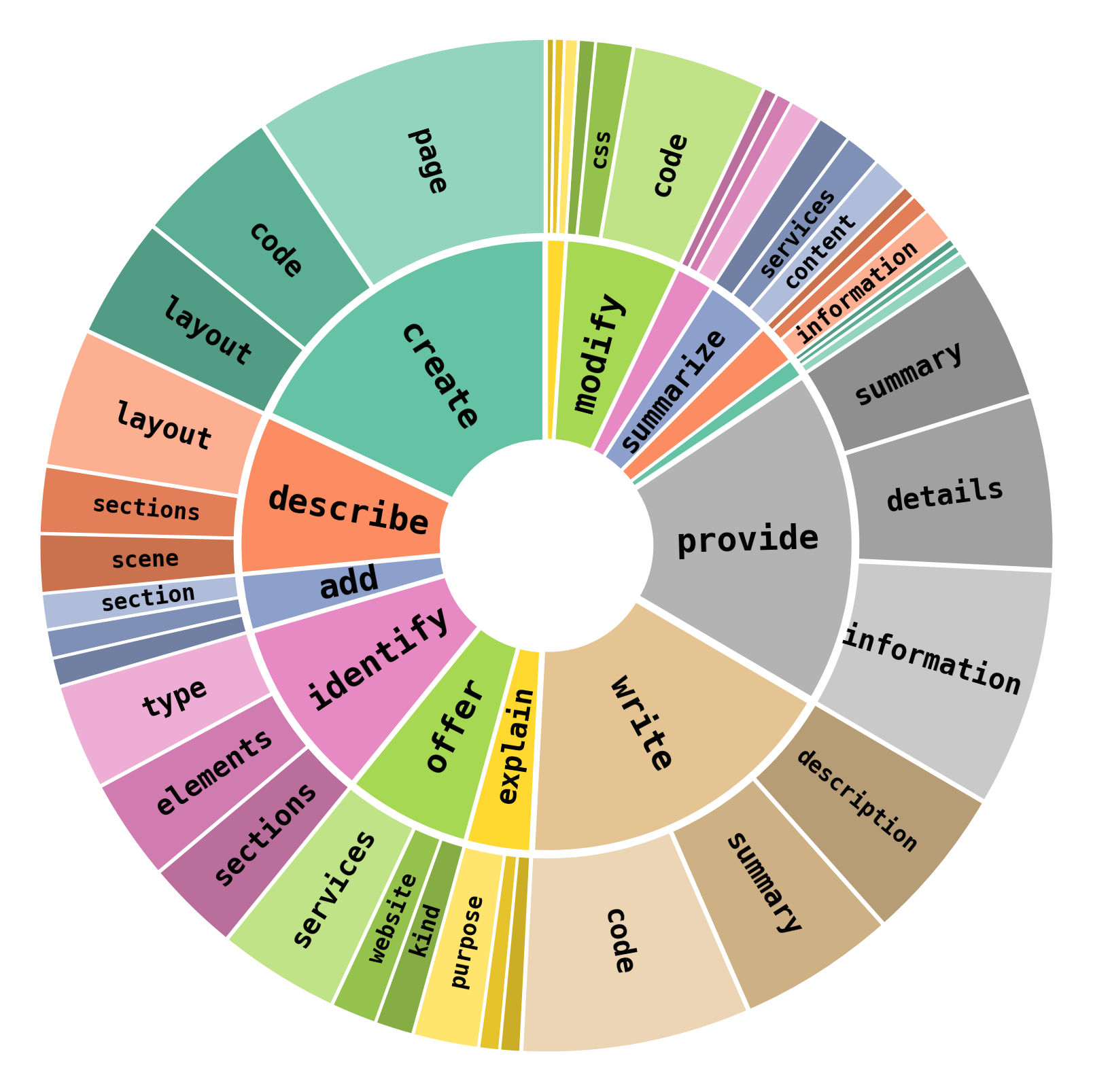}
      \end{minipage}
      \label{fig:indirect}
      }
    \caption{
          \textbf{Root verbs and top noun objects.} The charts show the most common root verbs and their top 3 noun objects in LLaVA-NeXT and \Method~data. Word combinations in LLaVA-NeXT data are quite concentrated, \eg, ``answer question'' and ``provide description''. Conversely, words in \Method~data are more natural and representative.} 
    \label{fig:word}
  \end{figure}

\begin{figure*}[t]
    \centering
    \includegraphics[width=1\linewidth]{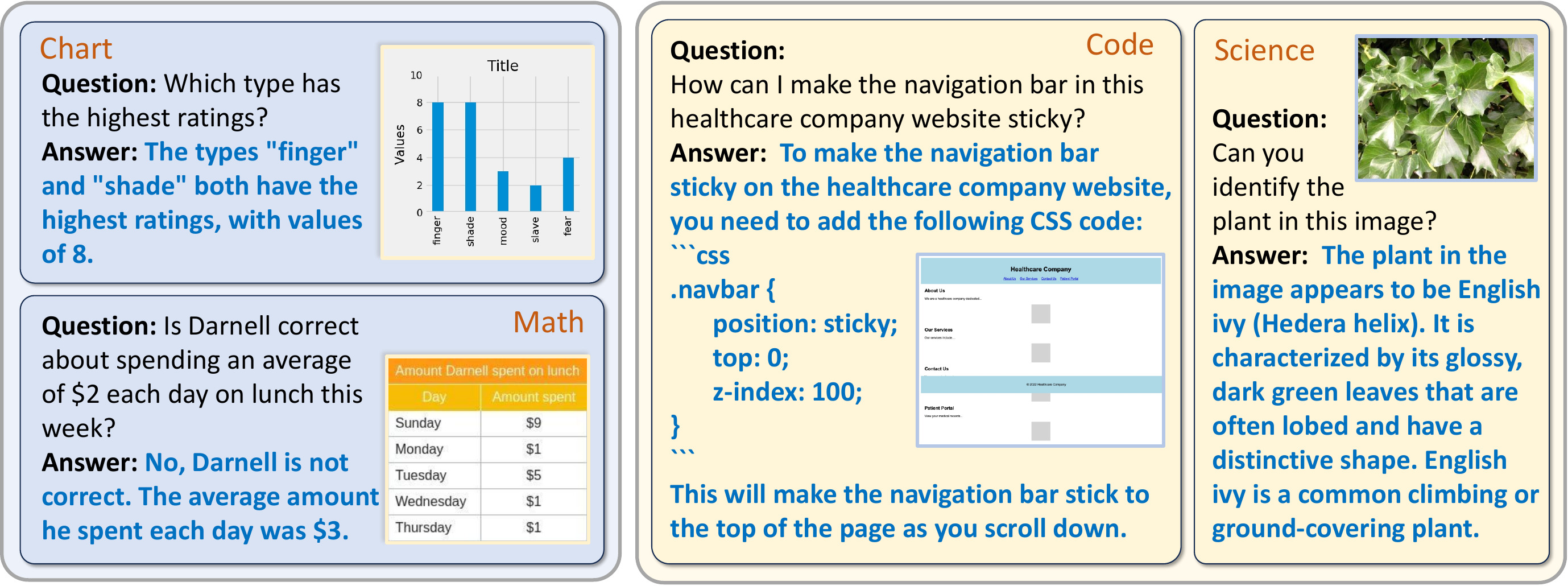}
    \caption{\textbf{\Method~synthetic data instances.} We present several examples illustrating the robustness of our data synthesis approach across diverse and rare domains~(\eg, Chart, Math, Code and Science). 
    Interestingly, the bottom-left figure shows that \Method~can generate complex math questions based on a chart image. \Method~identifies patterns and context within the table (\eg, daily spending amounts) and the related textual query ('Is Darnell correct about spending an average of \$2 each day on lunch?')
    }
    \label{fig:oasis_case}
  \end{figure*}

    \paragraph{\Method~data contains diverse language types.} Thank to the auto-regressive nature of our method, no language bias will be introduced by text prompt languages. Therefore, the generated instructions cover a wide range of language types. With the help of the langdetect library, we provide a visual breakdown of the language type distribution in \Method~and LLaVA-NeXT data, shown in~\cref{fig:language}. It is observed that our data contains more than 40 language types, while LLaVA-NeXT data only contains English. To our farthest knowledge, this is the first synthesized multi-modal dataset that covers such a wide range of languages.
  
  \paragraph{\Method~data covers a wide range of root verbs and top noun objects.} We provide a visual breakdown of the most common root verbs and top noun objects in \cref{fig:word}. Specifically, we extract the root verbs with a frequency over 1\% and their top 3 noun objects using the spaCy library. Compared to the LLaVA-NeXT data, the root verbs in \Method~data cover more natural and informative vocabularies. The top noun objects are also more diverse. Interestingly, LLaVA-NeXT data contains a lot of the combination ``answer question'', while \Method~has no such bias. 
  This suggests that \Method~captures a broader set of actions and interactions, reflecting a more nuanced and varied set of activities in the dataset. The heavy reliance on "answer questions" in LLaVA-NeXT hints at a potential overemphasis on QA tasks. More word-level analyses are in Appendix A.4.

  \paragraph{Cases of \Method~data.} We present cases of \Method~data in~\cref{fig:oasis_case}. It should be noted that our method can generate detailed and informative instructions based on the image topic. Additionally, we observe that our synthetic data covers a wide range of tasks, including object recognition, scene description, code understanding, etc. Such illustrations also support the aforementioned arguments about data diversity. More examples could be found in Appendix A.2.

\section{Experiments}
\label{sec:experiments}

\begin{table*}[t]
  \centering
  \caption{\textbf{Benchmarks used in the experiments and their corresponding domains.} We carefully choose 7 domains and 14 benchmarks to evaluate the effectiveness of our method comprehensively.}
  \small
  \begin{tabular}{cccccc}
    \toprule
     
  Domain & Benchmark & Domain & Benchmark & Domain & Benchmark \\
  \midrule
  \multirow{2}{*}{OCR} & TextVQA~\cite{singh2019towards} & \multirow{2}{*}{Science} & ScienceQA~\cite{scienceqa} & \multirow{2}{*}{General} & MMBench~\cite{mmbench}, MME~\cite{mme}\\
   & OCRBench~\cite{ocrbench}&  & AI2D~\cite{ai2d} & & MMVet~\cite{mmvet}, MMstar~\cite{mmstar}\\
    \midrule
  \multirow{2}{*}{Chart} & SeedBench2-Plus~\cite{seedbench,seedbench2_plus} & \multirow{2}{*}{Document} & DocVQA~\cite{docvqa} & Multi-discipline & MMMU-Pro~\cite{mmmu}  \\
   & ChartQA~\cite{chartqa} &  & InfoVQA~\cite{infovqa} & Visual Reasoning & GQA~\cite{gqa} \\
   \bottomrule
  \end{tabular}
  \label{tab:benchmarks}
  
  \vspace{-.5em}
  \end{table*}

 
\subsection{Implementation Details}

\paragraph{Synthesis details of \Method-500k data.} 
During the data synthesis process, we utilized two open-source models, Qwen2.5-VL-72B-Instruct as MLLM and Qwen2.5-72B-Instruct as LLM.
We randomly sample images from various public datasets as composed by the Cambrian-10M collection. Above models and dataset are easy to access via Huggingface, which guarantees the reproduction of this work.

\paragraph{Pre-training and supervised-finetuning data.} During the pre-training phase, we use LLaVA-Pretrain-558K~\cite{llava_next} for image-text alignment training. For supervised fine-tuning, we use officially open-source LLaVA-NeXT data as the baseline training set, where 15k instruction data from user data are not released. In the main experiment, we add our synthesized \Method~data to the original training set to evaluate the effectiveness of our method. In the ablation study, we prove that the multi-stage filtering mechanism improves the data quality.

\paragraph{Model.} For reproduction, we strictly follow the architecture of LLaVA-NeXT~\cite{llava_next}. The typical multi-modal large model consists of three key components: a visual encoder to extract visual representation, an image-text projector to align the visual and text modalities, and an LLM to generate the answer. We use CLIP-ViT-L as the backbone visual encoder and Vicuna-7B-v1.5 / Qwen2.5-7B-Instruct / Llama3-8B-Instruct as the LLM respectively. The projector is a 2-layer MLP with GELU activation function.

\paragraph{Training recipe.} Our experiments are conducted based mainly on the LLaVA-NeXT official codebase. We follow the 2-stage training strategy, including vision-language pre-training and visual instruction-tuning. In the pre-training stage, only the randomly initialized projector is trained. In the fine-tuning stage, the full parameters are tunable. 
For both pretrain and finetune stages, we use a cosine learning rate schedule with a warmup ratio of 0.03 and a global batch size of 128. The optimizer is AdamW~\cite{adamw} without weight decay. The learning rate is set to 1e-3 for pretrain and 1e-5 for finetune. It takes about 4 hours for pretraining and 30 hours for finetuning the baseline (LLaVA-NeXT).
\label{para:implementation}

\paragraph{Benchmarks.}
 To comprehensively evaluate the effectiveness of our evolutionary method, we select 14 benchmarks, with their sources and tested skills illustrated in \cref{tab:benchmarks}. These benchmarks encompass a wide range of vision-language tasks, including OCR, chart recognition, document analysis, general knowledge, multi-discipline, and visual reasoning. We evaluate the performance of our method on these benchmarks to demonstrate its effectiveness in improving the generalization ability of multi-modal models.

\subsection{Main Results}
\label{sec:main_results}

\begin{figure*}[t]
    \centering
    \subfloat[Comparison between baseline and \Method.]{
      \begin{minipage}[b]{0.27\textwidth}
        \includegraphics[width=1\textwidth]{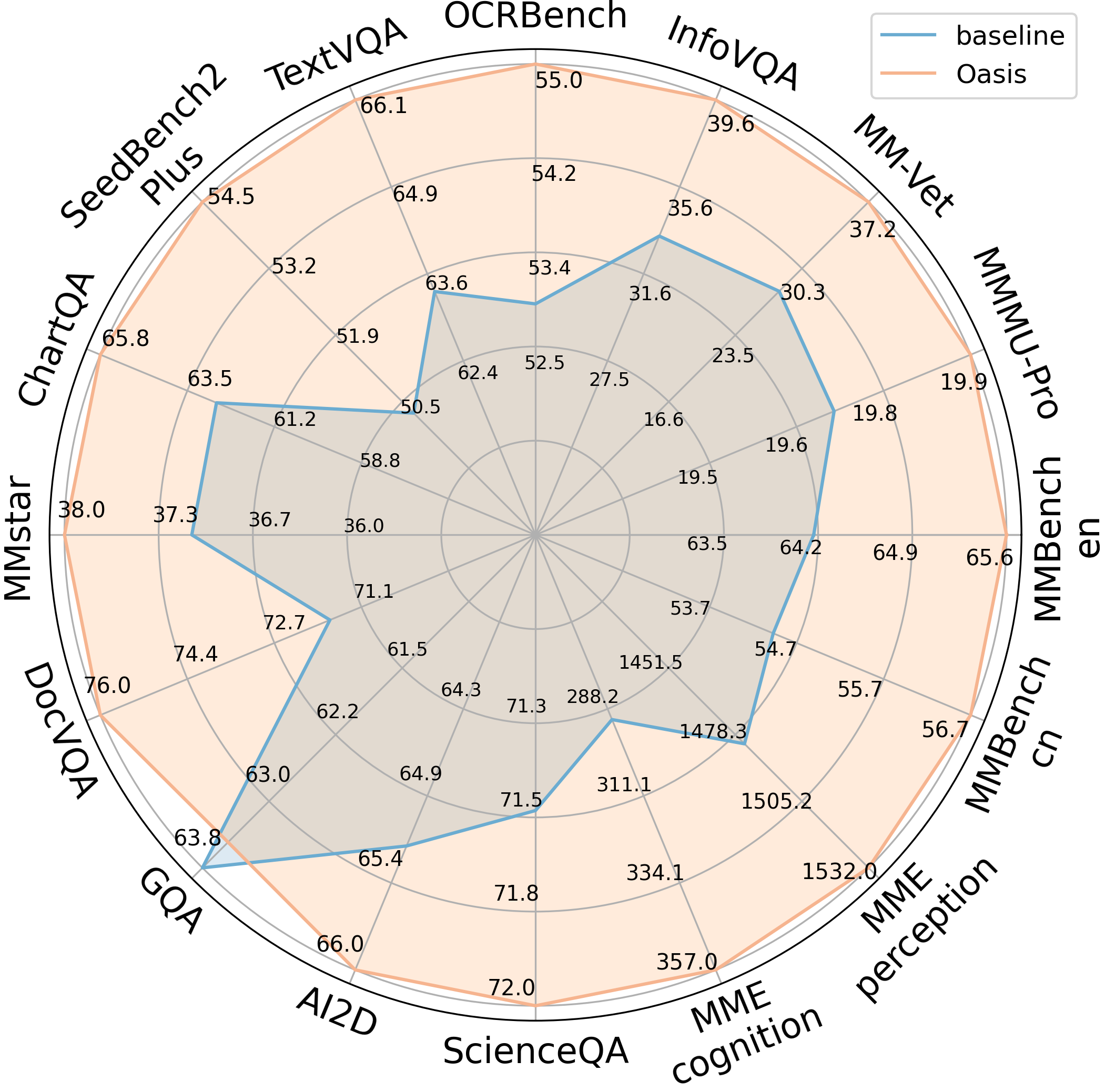}
      \end{minipage}
      \label{fig:radar}
      }
      \subfloat[OCR data generated by \Method.]{
      \begin{minipage}[b]{0.295\textwidth}
        \includegraphics[width=1\textwidth]{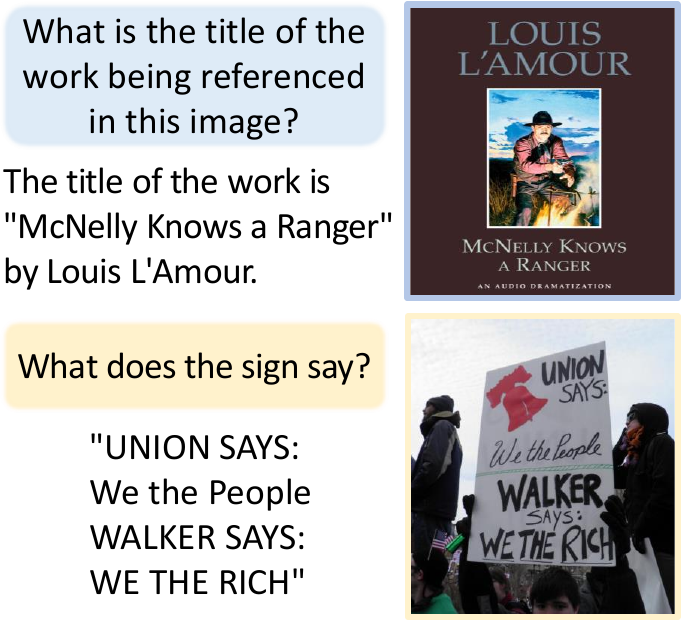}
      \end{minipage}
  \label{fig:ocr_case}
      }
      \subfloat[Instruct quality control accept/reject cases.]{
      \begin{minipage}[b]{0.387\textwidth}
        \includegraphics[width=1\textwidth]{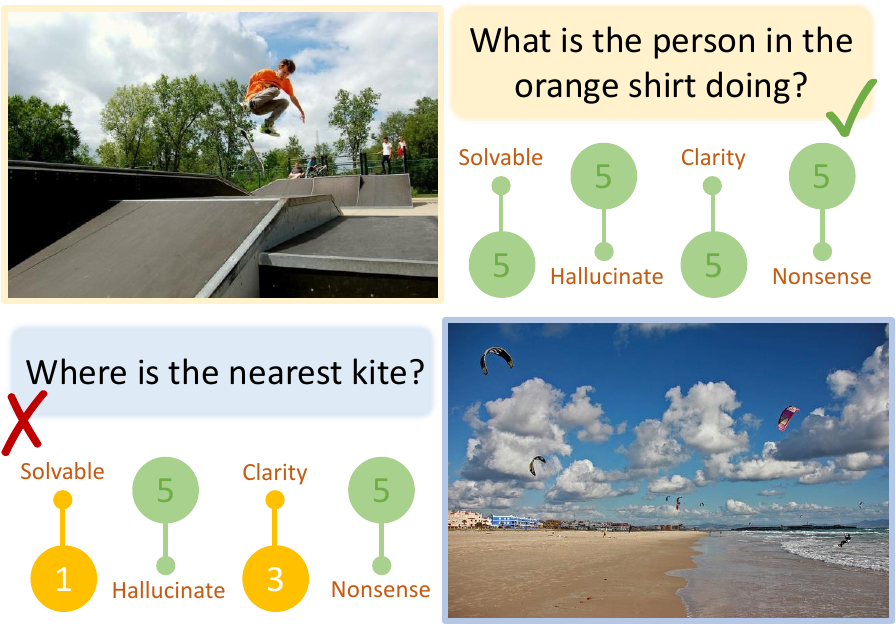}
      \end{minipage}
      \label{fig:inst_qc_case}
      }
    \caption{
          Figure(a) visualizes the improvements of \Method~over baseline, and significant overall enhancement can be observed. Figure(b) presents two OCR samples generated by \Method, revealing the data diversity. Figure(c) showcases the mechanism of instruction quality control. We accept the high-quality query above and reject the unsolvable and unclear query in the second case. 
          }
    \label{fig:r_o_qc}
    \vspace{-.5em}
\end{figure*}

\paragraph{Effectiveness of \Method.}  To demonstrate \Method's effectiveness, we perform a thorough evaluation of the MLLM trained with \Method~data across 14 benchmarks.
As shown clearly in~\cref{fig:radar}, our method provides a comprehensive and substantial improvement over the baseline. 
The detailed results are presented in \cref{tab:result}. 
We observe that our method consistently outperforms the baseline in almost all benchmarks, with an average improvement of 3.1\% / 1.8\% / 3.2\% for Vicuna1.5 / Qwen2.5 / Llama3 backbone, respectively. 
In particular, for Vicuna-7B-v1.5, improvements come from various domains, \eg, general knowledge, OCR, and document analysis. 
On the general knowledge domain, \Method~achieves 1.4\% and 2.3\% improvements over the baseline data on MMBench-en/cn. 
On OCR benchmarks, \Method~increases TextVQA and OCRBench for 2.7\% and 2.1\%.
On document analysis tasks, \Method~gains 4.3\% and 6.3\% over baseline. 
The above domains prove the diversity of our synthetic data, and the above improvements reveal the effectiveness of our method in enhancing the generalization ability of MLLMs.

\paragraph{Comparison with other synthesis methods.} To further validate the efficiency of our method, we compare \Method~data with four other datasets of equivalent size in Vicuna-7B-v1.5.
\underline{(1)} The original annotation of \textbf{Oasis}~images in Cambrian-10M~\cite{cambrian}.
\underline{(2)} Upsampled LLaVA SFT data. 
\underline{(3)} MMEvol data. 
\underline{(4)} DenseFusion-1M data~\cite{densefusion} (combined with random detailed description instructions in LLaVA paper). 
We sample each dataset to the same size with \Method~for fair comparison.
The results in~\cref{tab:result}~show that~\Method~data outperforms other methods remarkably. 
OCR-related tasks benefit the most from our data, with a steady improvement of more than 2\% across DocVQA, InfoVQA, TextVQA and OCRBench over most methods. 
On general tasks, \Method~data also shows an overall advantage on MME, MMstar and MM-Vet. 
We argue that \Method~data is inherently easier for the model to learn thanks to its prompt-free nature, as the generated data adheres to the internal knowledge distribution of MLLMs.

\begin{table*}[t]
  \caption{\textbf{Main results.} All baseline experiments are reproduced with LLaVA-NeXT's official code. \Method~introduces extra synthetic data to baseline in the fine-tuning part. 
  We observe consistent performance gains across various benchmarks, including general and complex tasks. 
  The improvement \textbf{+3.1\% / +1.8\% / +3.2\%} is particularly notable in average, indicating that the synthetic data effectively supplements real-world data.
  Our data also outperforms other synthetic data in most benchmarks, with overall improvement ranging from 1.2\% to 4.5\%.  
  The last part shows the efficiency of scaling \Method~data, with 5.2\% improvement in average.
  }
  \begin{adjustbox}{width=\textwidth,center}
  \begin{tabular}{ccc@{\hspace{1.5\tabcolsep}}c@{\hspace{1.7\tabcolsep}}c@{\hspace{0.9\tabcolsep}}c@{\hspace{1.7\tabcolsep}}ccccc@{\hskip 10pt}c@{\hspace{1.2\tabcolsep}}c@{\hskip 10pt}cc@{\hspace{1.5\tabcolsep}}c}
    \toprule
  Method & MMBench & MME & MMStar & MMVet & MMMU$^{\text{pro}}_{\text{std}}$ & GQA & AI2D & Sci & Doc & Info & Chart & Seed2 & Text & OCR & AVG.\\
  \midrule
  
  \textit{Baseline}-Vicuna1.5 & 64.2 / 54.4 & 1482 / 291 & 37.1 & 28.0 & 19.7 & 63.8 & 65.2 & 71.5 & 71.7 & 33.3 & 62.7 & 50.3 & 63.4 & 52.9 & 53.0\\
  
  LLaVA (upsample) & 64.8 / 54.9 & 1461 / 353 & 37.6 & 34.3 & 19.8 & 63.9 & 65.9 & 71.6 & 67.8 & 29.4 & 64.4 & 51.3 & 64.0 & 52.6 & 53.7\\
  Densefusion & \textbf{67.4} / 56.2 & 1523 / 333 & 37.8 & 30.2 & 19.4 & 63.9 & 65.4 & 71.9 & 69.2 & 32.9 & 61.5 & 53.6 & 65.4 & \textbf{55.4}  & 54.3\\
  Cambrian & 66.8 / 56.6 & 1504 / 329 & 37.8 & 32.4 & 19.7 & \textbf{64.1} & \textbf{69.4} & 70.6 & 73.8 & 37.7 & 63.4 & 53.9 & 63.7 & 52.3 & 54.9\\
  MMEvol & 63.6 / 53.8 & 1503 / 316 & 32.3 & 34.9 & 19.1 & 63.4 & 64.9 & 54.4 & 64.7 & 30.5 & 61.5 & 53.8 & 62.8 & 51.7 & 51.6\\
  \rowcolor{green!15}
  \Method-Vicuna1.5 & 65.6 / \textbf{56.7} & \textbf{1532 / 357} & \textbf{38.0} & \textbf{37.2} & \textbf{19.9} & 63.5 & 66.0 & \textbf{72.0} & \textbf{76.0} & \textbf{39.6} & \textbf{65.8} &  \textbf{54.5} & \textbf{66.1} & 55.0 & \textbf{56.1}\\
  \midrule

  \textit{Baseline}-Qwen2.5 & 76.0 / 74.4 & 1577 / 405 & 51.2 & 32.2 & 28.7 & \textbf{64.1} & 76.7 & \textbf{83.1} & 74.0 & 35.9 & 73.1 & 62.6 & 65.0 & 56.0 & 61.4 \\
  \rowcolor{green!15}
  \Method-Qwen2.5 & \textbf{77.4} / \textbf{74.5} & \textbf{1598 / 439} & \textbf{51.3} & \textbf{35.2} & \textbf{29.5} & 64.0 & \textbf{78.7} & 83.0 & \textbf{77.3} & \textbf{41.6} & \textbf{74.3} & \textbf{65.1}& \textbf{65.9} & \textbf{58.2} & \textbf{63.2} \\
  \midrule
  
  \textit{Baseline}-Llama3 & 71.4 / 66.6 & \textbf{1539} / 313 & 45.6 & 29.6 & 21.7 & 63.8 & 72.1 & 80.1 & 65.8 & 28.3 & 64.4 & 54.8 & 61.1 & 51.2 & 55.8 \\
  \rowcolor{green!15}
  \Method-Llama3 & \textbf{72.7} / \textbf{67.5} & 1522 / \textbf{355} & \textbf{46.4} & \textbf{36.6} & \textbf{22.4} & \textbf{63.9} & \textbf{74.6} & \textbf{81.0} & \textbf{73.0} & \textbf{34.7} & \textbf{71.2} & \textbf{59.5} & \textbf{64.9} & \textbf{55.8} & \textbf{59.0} \\
  \midrule[.8pt]
  
  LLaVA-100k & 57.6 / 46.9  &  1408 / 284  &  37.7  &  26.7  &  18.6  &  57.9  &  56.3  &  \textbf{70.8}  &  52.0  &  25.3  &  47.3  &  44.9  &  56.4  &  40.0  &  46.5 \\
  \quad +\Method-150k  & 55.9 / 48.2  &  1312 / 297  &  35.0  &  29.7  &  17.7  &  58.4  &  56.8  &  67.7  &  52.5  &  25.8  &  50.2  &  45.7  &  57.2  &  42.4  &  46.6$_\text{\textcolor{darkgreen}{+0.1}}$ \\
  \quad +\Method-300k  & 57.5 / 49.0 & 1419 / 294 & 37.3 & 31.3 & 18.8 & 58.0 & 56.8 & 68.3 & 53.8 & 25.7 & 51.5 & 45.4 & 57.7 & 44.9 & 47.7$_\text{\textcolor{darkgreen}{+1.2}}$ \\
  \rowcolor{green!15}
  \quad +\Method-500k & \textbf{58.8 / 51.4} & \textbf{1448 / 331} & \textbf{40.5} & \textbf{38.3} & \textbf{19.1} & \textbf{58.9} & \textbf{58.0} & 70.3 & \textbf{63.9} & \textbf{31.7} & \textbf{57.6} & \textbf{52.3} & \textbf{61.9} & \textbf{50.9} & \textbf{51.7}$_\text{\textcolor{darkgreen}{+5.2}}$ \\
  \bottomrule
  
  \end{tabular}
  \end{adjustbox}
  \label{tab:result}
  
\end{table*}

\paragraph{Scaling synthetic data.} We conduct a scaling experiment to investigate the impact of the amount of synthetic data on the performance of MLLM. 
Specifically, we respectively add 150k, 300k, and 500k \Method~data to 100k LLaVA-NeXT SFT data, and compare the performance of these models. 
We downsample LLaVA-NeXT data to 100k to avoid the common problem of `data mixture ratio' and thus fully reveal our data efficiency, but keep the basic model capacity simultaneously.
\Cref{tab:result}~shows that when the size of the data increases from 0 to 500k, the overall performance of the model improves steadily. 
After incorporating 500K \Method~data, the average score improved by 5.2\%, providing strong evidence of the effectiveness of our data. 
Additionally, it is noteworthy that an increase from 300K to 500K data still results in a substantial 4.0\% improvement, indicating that our scaling remains effective even in large data amounts and enables the model to achieve consistent gains.
We argue that when scaled up, our synthetic data can continuously inject knowledge into the model, progressively enhancing its capabilities.



\begin{table}[t]
  \centering

  \caption{\textbf{OCR domain experiments.} Adding specific-domain data generated by \Method~contributes to consistent improvements to various OCR-related benchmarks.}
  \resizebox{.47\textwidth}{!}{
  \begin{tabular}{lcccccc @{\hskip 10pt} c}
    \toprule
   OCR Data & Doc & Text  & Chart & OCR & Info & AI2D & Seed2\\
   \midrule
   No & 71.7 & 63.4 & 62.7 & 52.9 & 33.3 & 65.2 & 50.3 \\
   OCR-30k & 73.5 & 63.7 & 64.5 & 54.5 & 35.6 & 65.1 & \textbf{52.6} \\
   OCR-70k & \textbf{75.0} & \textbf{64.7} & \textbf{66.2} & \textbf{55.3} &\textbf{ 37.1} & \textbf{66.3} & 52.4 \\
  \bottomrule
  \end{tabular}}
  \label{tab:ocr_exp}
  \vspace{-1em}
\end{table}

\subsection{Specific-domain Data Synthesis}
\Method~is a flexible method for the synthesis of data in a specific domain, since the input image inherently sets the data domain. By conditioning on the characteristics of the input image, \Method~effectively generates domain-specific attributes, which makes it possible to train models in fields where data might be scarce. We take ``OCR'' as a typical data domain to validate the effectiveness of \Method in this section. Additionally,  Appendix A.5 includes an application of \Method~in the medical domain.

\paragraph{OCR data source and the synthetic data.} We select 311k images from 24 OCR-related datasets, and generate data with \Method. Filtered with data categorization and quality control, 70k data remains. We present several examples in~\cref{fig:ocr_case}, showing the diversity of \Method~data extends beyond basic OCR tasks.
It can incorporate a broad spectrum of question types designed to test various reasoning and comprehension abilities. \Method~creates questions that not only require direct text extraction but also challenge the model’s capacity for contextual understanding, attribute-based reasoning, and logical deduction.
More discussions and examples can be found in Appendix A.2. 
\vspace{-.5em}



\paragraph{Effectiveness of the synthetic OCR data.} As empirical results presented in \cref{tab:ocr_exp}, the synthetic OCR data demonstrates notable efficacy in improving model performance. 
We select 7 OCR-related or text-rich benchmarks and find that our OCR-oriented synthesis data could provide steady improvement to these tasks. 
By introducing additional synthetic data, the model gains exposure to a broader range of problem types, enhancing its ability to generalize to previously unseen tasks. 
In particular, the synthetic data serves to bridge gaps in the distribution of real-world data, covering edge cases and rare patterns that might otherwise be underrepresented. Consequently, this blend of real and synthetic data results in a more robust model that performs reliably across challenging OCR problems.

\begin{table*}[t]
  \caption{\textbf{Ablation results.}  
    The \textbf{caption data recycling}, \textbf{instruction quality control} and \textbf{response quality control} experiments respectively. 
    The caption recycling result provides evidence that we can easily reuse the waste caption data for further improvement. 
    The instruction quality control mechanism resulted in a 1\% improvement, proving to be both effective and indispensable.
    Response quality control methods fail to provide any additional enhancements and are therefore unnecessary.
  }
  \vspace{-.5em}
  \resizebox{\textwidth}{!}{
      \begin{tabular}{c@{\hspace{1.7\tabcolsep}}ccc@{\hspace{1.7\tabcolsep}}c@{\hspace{0.9\tabcolsep}}c@{\hspace{1.7\tabcolsep}}ccccc@{\hskip 10pt}c@{\hspace{1.2\tabcolsep}}c@{\hskip 10pt}ccc}
        \toprule
      Method & MMBench & MME & MMStar & MMVet & MMMU$^{\text{pro}}_{\text{std}}$ & GQA & AI2D & Sci & Doc & Info & Chart & Seed2 & Text & OCR & AVG.\\
      \midrule
    
      \rowcolor{gray!15}
      w/o caption & 65.6 / \textbf{56.7} & \textbf{1532} / 357 & 38.0 & \textbf{37.2} & 19.9 & 63.5 & \textbf{66.0} & 72.0 & 76.0 & 39.6 & 65.8 &  54.5 & 66.1 & 55.0 & 56.1\\
      + 250k caption & \textbf{65.8} / 54.6 & 1496 / \textbf{368} & \textbf{40.2} & 35.9 & \textbf{20.8} & \textbf{64.0} & 65.7 & \textbf{72.5} & \textbf{77.3} & \textbf{40.5} & \textbf{66.0} & \textbf{55.3} & \textbf{66.3} & \textbf{56.1} & \textbf{56.4} \\
      \midrule
    
      \rowcolor{gray!15}
      w/ Inst. QC & 63.8 / \textbf{55.1} & \textbf{1530} / \textbf{315} & 37.9 & \textbf{36.8} & \textbf{19.3} & \textbf{64.0} & 65.2 & \textbf{71.8} & \textbf{74.8} & \textbf{38.7} & \textbf{64.3} & 53.7 & \textbf{64.5} & 52.1 & \textbf{54.9} \\
      w/o Inst. QC & \textbf{64.7} / 52.9 & 1516 / 308 & \textbf{39.8} & 35.1 & 19.0 & 63.9 & \textbf{65.4} & 71.4 & 67.7 & 31.4 & 63.8 & \textbf{53.8} & 63.6 & \textbf{55.1} &53.9 \\
    
      \midrule
      \rowcolor{gray!15}
      w/o Resp. QC & 65.6 / \textbf{56.7} & 1532 / \textbf{357} & 38.0 & 37.2 & 19.9 & 63.5 & 66.0 & 72.0 & \textbf{76.0} & \textbf{39.6} & 65.8 &  \textbf{54.5} & \textbf{66.1} & 55.0 & \textbf{56.1}\\
      NLL sample  & \textbf{66.1} / 56.2 & \textbf{1535} / 323 & \textbf{39.2} & 40.3 & 20.2 & \textbf{64.0} & \textbf{66.3} & \textbf{72.8} & 69.8 & 32.8 & \textbf{66.1} & 53.4 & 66.0 & \textbf{56.8} & 55.4 \\
      Scoring  & 56.7 / 54.9 & 1522 / 288 & 36.9 & \textbf{43.6} & \textbf{20.5} & 63.3 & 65.5 & 72.6 & 68.9 & 32.4 & 64.5 & 53.1 & 64.8 & 53.7 & 54.5 \\
    
      \bottomrule
      
      \end{tabular}
  }
  \label{tab:ablation}
  \vspace{-1.em}
  \end{table*}

\subsection{Ablation Studies}
\Method~involves two essential synthesis steps, \ie, data categorization and instruction quality control. In the following, we discuss the recycling of the filtered-out data in data categorization and the effects of instruction quality control. Then, we ablate the response quality control is unnecessary.

\vspace{-.5em}
\paragraph{Recycling of caption data.} 
As stated in~\cref{para:categorization}, we utilize an LLM to process the data obtained from the first step, where the captions are removed while the instructions are retained.
Given that the pass rate is only 49.90\%, approximately half of the caption-like data remains unused. 
Therefore, we explore strategies to recycle these data.
Since some data contain mixed special tokens or irrelevant fields, we first apply rule-based filtering for an initial screening. 
Then, we further refine the data using Qwen2.5-72B-Instruct to remove entries that are unsuitable as captions. 
We apply few-shot in this phase for better accuracy.
Eventually, 250k high-quality captions remain.
We combine these captions with a random instruction for detailed image description listed in LLaVA paper and add them into the \Method-data.
The data in~\cref{tab:ablation}~suggest that this data yields promising results, with 12 out of 16 metrics surpassing the baseline. 
This result effectively demonstrates that we can efficiently recycle the waste data at an extremely low cost.

\vspace{-.5em}
\paragraph{Effects of instruction quality control.} After data categorization, instruction quality control is applied to filter out low-quality instructions in four dimensions: solvability, clarity, hallucination, and nonsense. Previous works
have adequately verified the reliability of LLM/MLLM-as-a-judge for their high human agreement~\cite{llmjudging, mllmjudging}. We further conduct an ablation study to evaluate the impact of this quality control mechanism on the data quality and model performance. In particular, we compare the performance of the model trained with and without quality-controlled 200k data, respectively. Specifically, The acceptance rate of high-quality instructions is a reasonable 50.90\%. Therefore, the 200k data without quality control is expected to contain 100k ``low-quality" instructions. \Cref{tab:ablation}~presents the results. With the quality control mechanism, the model achieves a notable 1\% overall enhancement. In both DocVQA and InfoVQA, the model remarkably gains over 7\% improvement. This result demonstrates the necessary role of data quality control in \Method.

\vspace{-.5em}
\paragraph{Attempts on response quality control.}
\label{para:resp_qc}
To study the necessity of response quality control, we try 2 methods to filter out low-quality responses. 
\underline{(1)} NLL reject sampling: We sample 5 responses for each instruction and calculate their average token NLL (negative log likelihood), and keep the highest one as the final response, which is the answer with the most confidence~\cite{simpo}.
\underline{(2)} MLLM scoring: We score responses from 3 dimensions (helpfulness, truthfulness and instruction-following) on a scale of 1 to 5 with Qwen2-VL-72B-Instruct, and filter out responses that do not get full marks. 
Results in~\cref{tab:ablation}~reveal that these response quality controls are either ineffective or even harmful. The average score drops by 0.7\% and 1.6\% in the 2 scenarios.
We argue that high-quality instructions could inherently derive good responses from SOTA MLLMs. Unexpected biases would be introduced by extra filtering procedures.


\section{Conclusions}
\vspace{-.5em}
Based on the proposed method for synthesizing multi-modal training data, we have demonstrated that it is possible to generate high-quality and diverse multi-modal data using only images, without compromising data quality. Through a careful quality control process and extensive experimentation, we have shown that \Method~significantly enhances the performance of MLLM with various backbones, and outperforms other synthesis methods by a large margin. Our approach offers a promising solution to the challenges posed by the unavailability and high cost of traditional multi-modal data synthesis. It also opens new possibilities to improve MLLMs in specific domains. In conclusion, \Method~not only addresses the scarcity of multi-modal data but also provides a scalable way to improve the capabilities of MLLM, paving the way for more efficient and accessible training paradigms. We will release our code and dataset to help the future research of the community.



{
    \small
    \bibliographystyle{ieeenat_fullname}
    \bibliography{main}
}

\clearpage
\appendix
\onecolumn


\section{Additional Experiments and Discussions}

\subsection{Relation to Magpie}

Magpie is a framework for efficient LLM training data synthesis that leverages a similar ``hooking'' method like ours. In LLM scenarios, Magpie divides the input to an LLM into three parts: the pre-query template, the query, and the post-query template. For Llama-3-8B-Instruct, an input can be ``\texttt{\textcolor{blue!80}{<|start\_header\_id|>user <|end\_header\_id|>}\textcolor{red!80}{Hi! <|start\_header\_id|> assistant<|end\_header\_id|>}}". They feed only the pre-query template (\textcolor{blue!80}{blue} part) into Llama-3-8B-Instruct and extract potential instruction output, leading to 300K high-quality and diverse instances, which could be further extended easily. They use their collected data to fine-tune Llama-3 and achieve remarkable advantages against six other state-of-the-art open-source instruction tuning datasets.

We gain our inspiration from Magpie and extend the method to multimodal scenarios. Consequently, we explore the feasibility of this idea to synthesize multimodal data in depth and propose a novel method, \Method, which could lead to huge improvements in MLLMs.
Compared to Magpie, a primary difference in our method is that we include the image as an additional input, which allows the MLLM to generate instruction based not only on its internal knowledge but also on the visual information. This attribute enables the image domain to control the synthesized instruction domain, making our method more versatile and applicable to a broader range of tasks. It is also worth noting that we handcraft all-around quality control means specifically for multimodal data, which effectively ensures the quality of the synthesized data and paves the way for the community to explore better multimodal data synthesis.

\subsection{Cases of \Method-500k}

\begin{figure}[t]
    \centering
    \includegraphics[width=.9\linewidth]{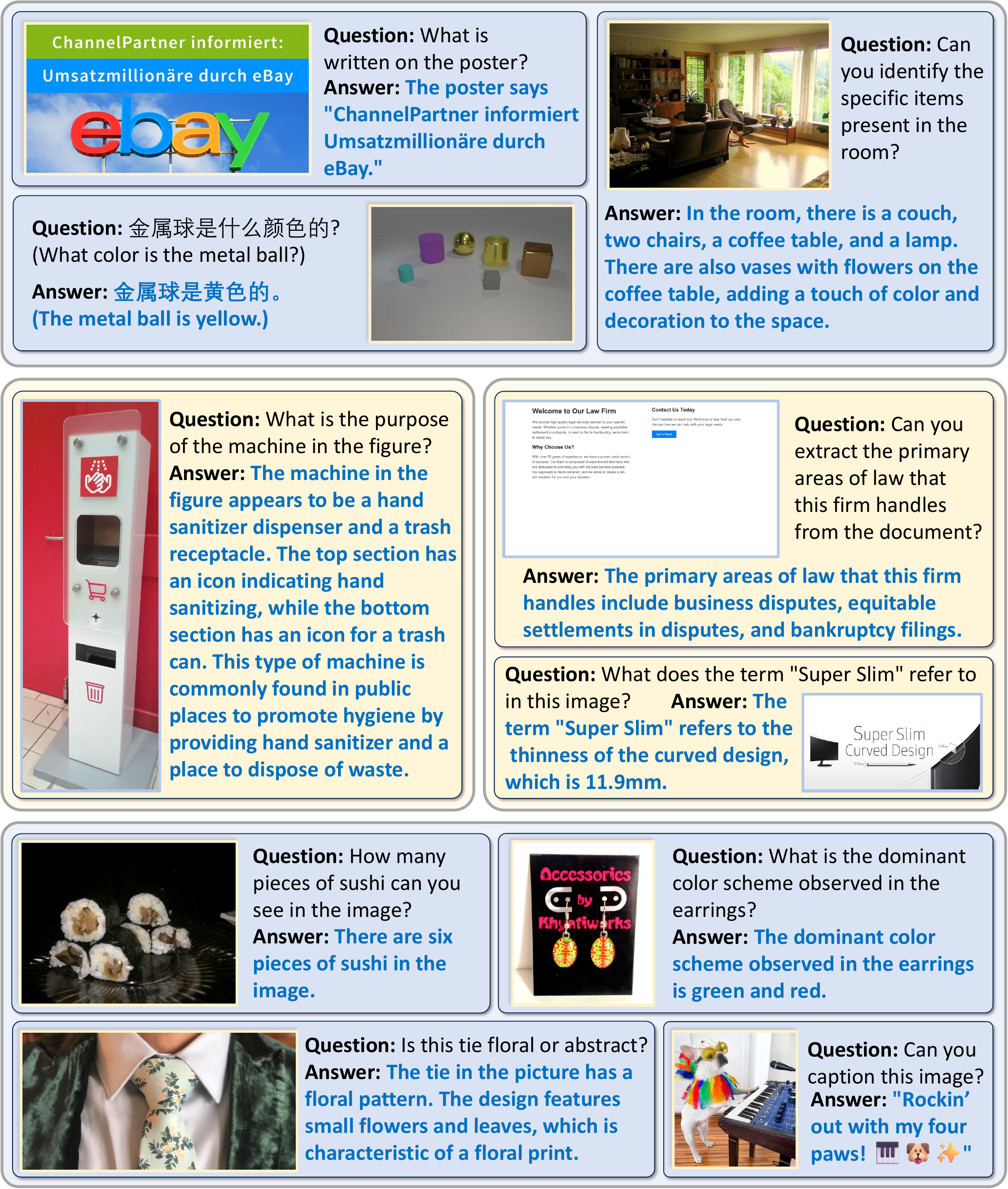}
    \vspace{-.5em}
    \caption{\textbf{\Method~data cases.} This figure shows several cases of \Method~data. It can be observed that the data synthesized by \Method~is diverse and creative, covering a wide range of tasks and domains.}
    \vspace{-.5em}
    \label{fig:supp_case}
\end{figure}

We provide more cases of \Method-500k data in~\cref{fig:supp_case}. It can be observed that the generated instruction encompasses a wide range of tasks and domains, including OCR, object recognition, scenario understanding, commonsense knowledge, etc. Thanks to the `hooking' method, the instruction is diverse, creative, and enlightening, which is beneficial for the multimodal model to extend its generalization ability.

\subsection{Instruction Quality Control Details}
In Step 3 of our method, a comprehensive quality control process is conducted to ensure the quality of the synthesized data. In detail, we evaluate the solvability, hallucination, clarity, and nonsense of the instruction and filter out ~50\% of the data. Here we provide the detailed filtering criteria for each dimension. Each dimension of instruction is scored on a scale of 1 to 5, with 1 being the worst and 5 being the best. For hallucination and nonsense, we only retain the data with a score of 5, since the existence of any hallucination or nonsense could lead to misleading training, harming the model's performance and generalization ability. For solvability and clarity, only the data that satisfies each score being greater than or equal to 3, and the sum of the scores being greater than or equal to 7 will be retained. This standard is set as a balance of filterability, synthesis efficiency, and data diversity.

\begin{figure}[t]
    \centering
    \subfloat[Solvability Score]{
      \begin{minipage}[b]{0.22\textwidth}
        \includegraphics[width=1\textwidth]{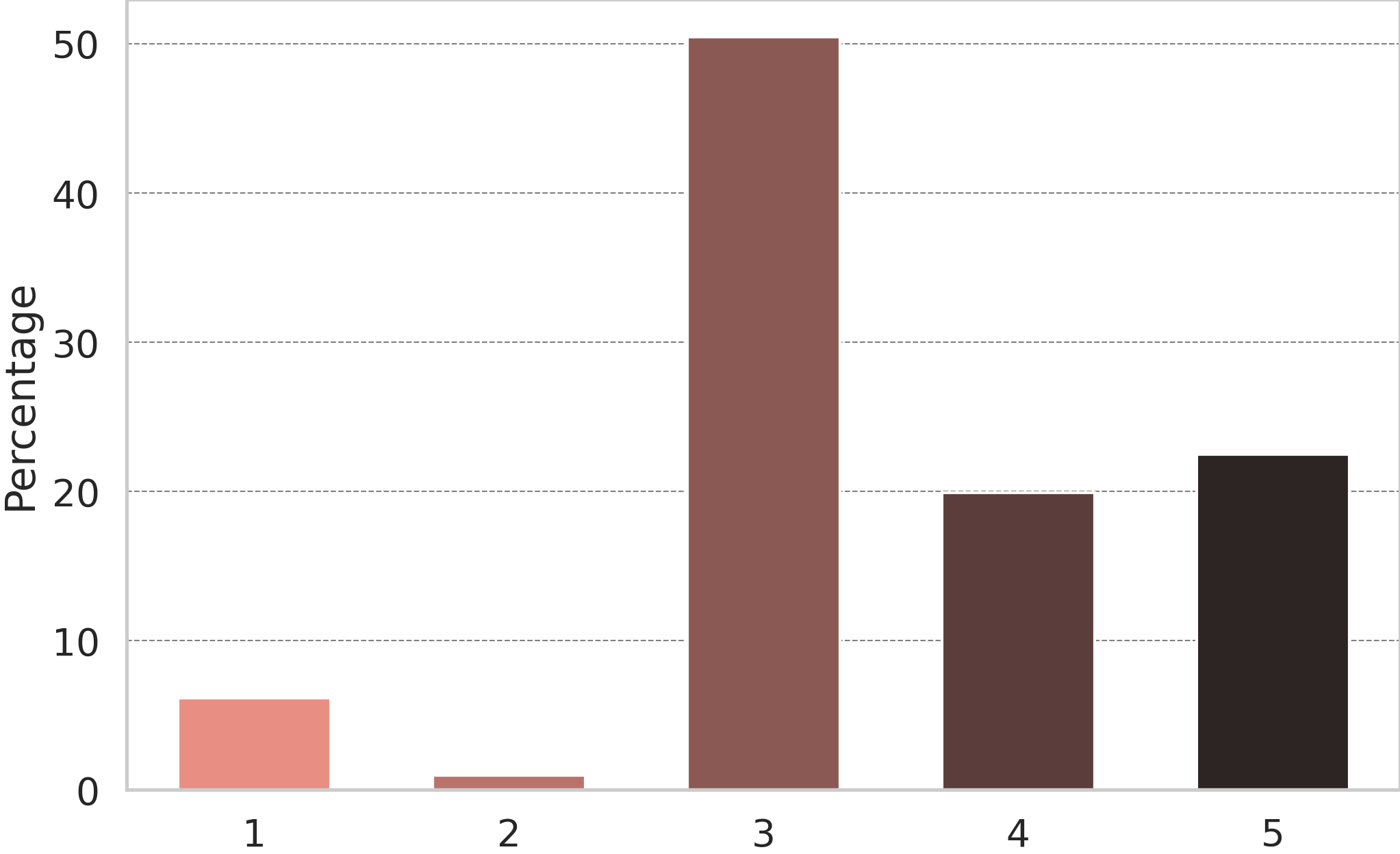}
      \end{minipage}
      \label{fig:direct}
      }
      \subfloat[Clarity Score]{
      \begin{minipage}[b]{0.22\textwidth}
        \includegraphics[width=1\textwidth]{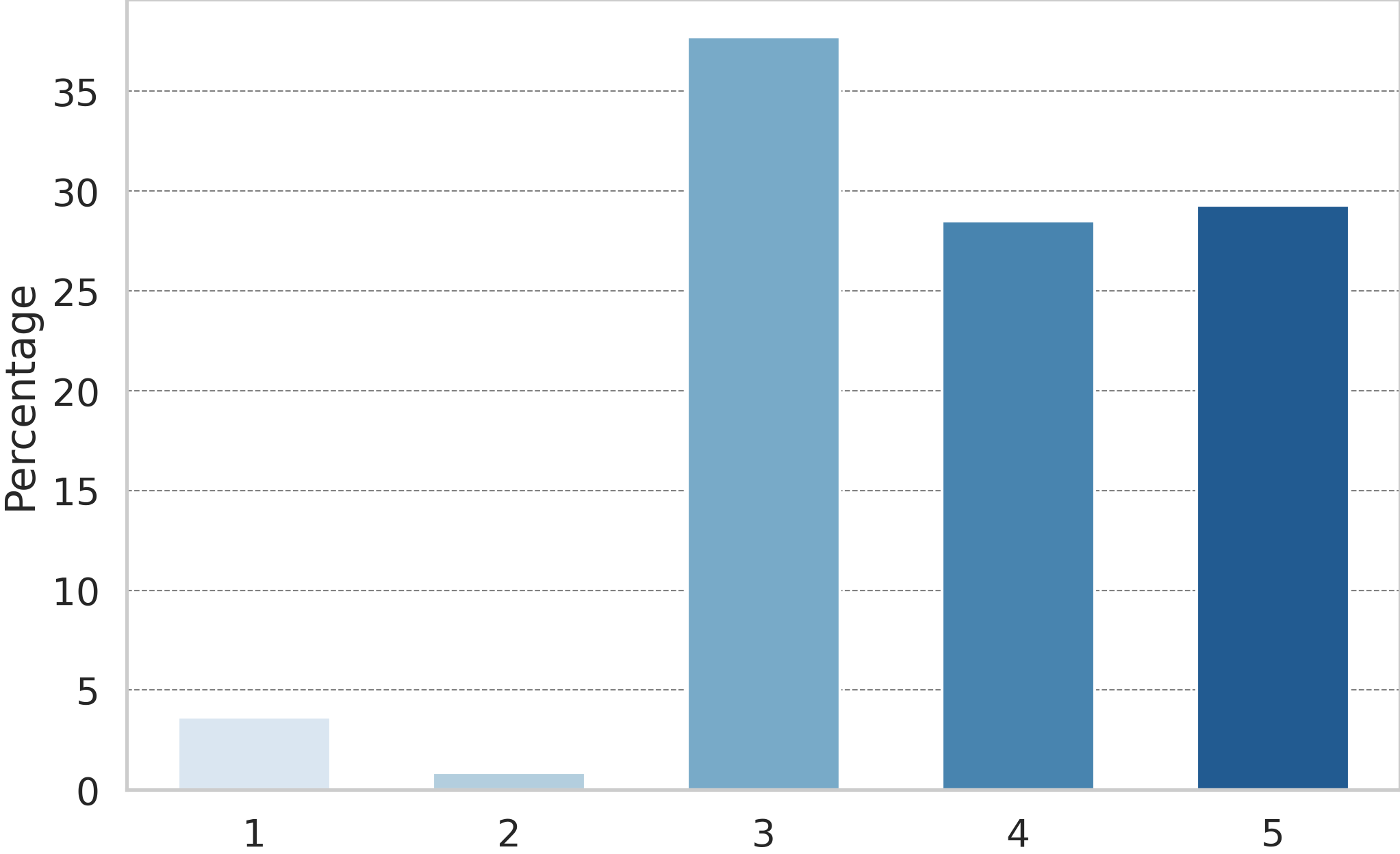}
      \end{minipage}
  \label{fig:indirect}
      }
      \subfloat[Hallucination Score]{
      \begin{minipage}[b]{0.22\textwidth}
        \includegraphics[width=1\textwidth]{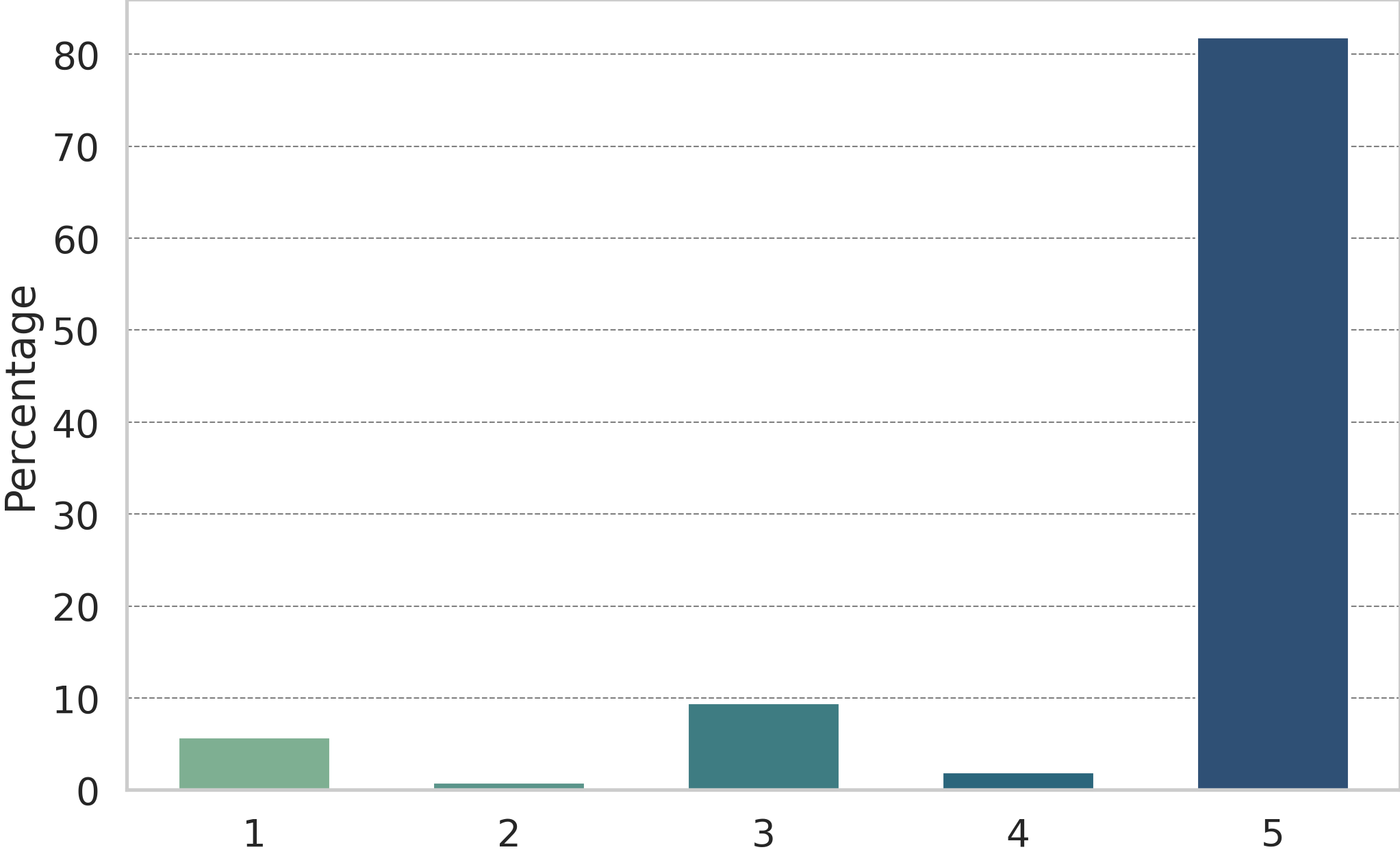}
      \end{minipage}
      \label{fig:direct}
      }
      \subfloat[Nonsense Score]{
      \begin{minipage}[b]{0.22\textwidth}
        \includegraphics[width=1\textwidth]{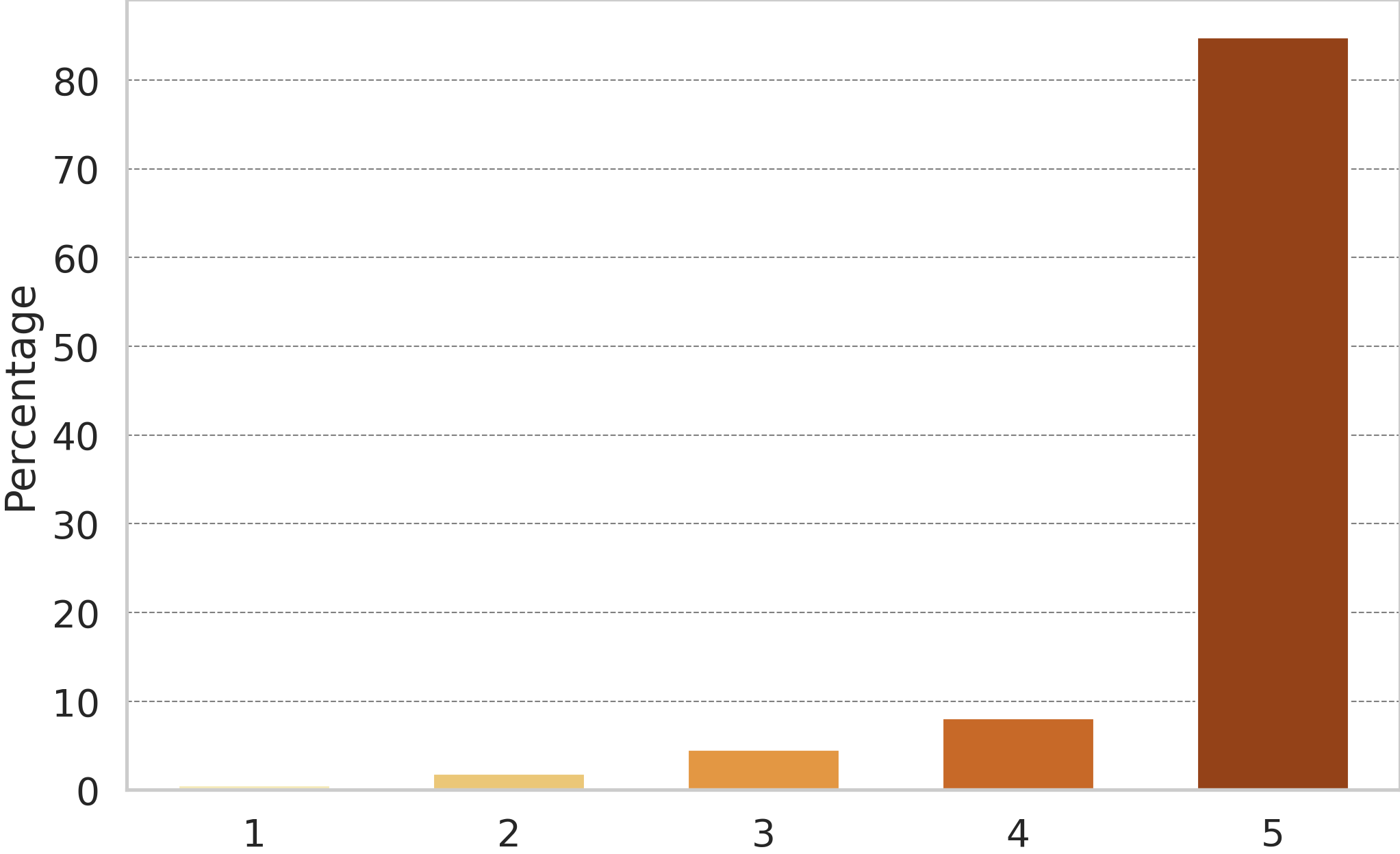}
      \end{minipage}
  \label{fig:indirect}
      }
    \caption{
          \textbf{Distribution of instruction quality scores.} We use MLLM to evaluate the solvability, clarity, and hallucination scores, and LLM to evaluate the nonsense score. The majority of instructions contain no hallucination and nonsense, while the solvability and clarity attributes are more evenly distributed.}
    \label{fig:score}
  \end{figure}

\vspace{-1em}
\paragraph{Distribution of instruction quality scores.} \Cref{fig:score} here illustrates the range and distribution of quality scores assigned to instructions. The results show that the majority of instructions are rated highly in terms of hallucination and nonsense, which can be attributed to the strength of MLLM. In comparison, the solvability and clarity scores are more evenly distributed, which leads to a sufficient filtering mechanism.

\subsection{More data analysis}

\paragraph{\Method~data has large type-token ratios.} We calculate the type-token ratio (TTR) of the instruction and response data. The TTR is defined as the ratio of the number of unique words to the total number of words in the dataset. As shown in \cref{tab:ttr}, the TTR of \Method~is significantly higher than that of LLaVA-NeXT, especially in the instruction data. This indicates that \Method~data is more lexically diverse and covers a wider range of topics, which can help improve the generalization ability.

\subsection{Application on medical area}

We validate~\Method~on medical benchmarks in~\cref{tab:medical}. We sample 15k images from the MedTrinity-25M dataset and create 2k medical training data with~\Method. 
We SFT the LLaVA-NeXT baseline with 4k sampled LLaVA data and 2k LLaVA data + 2k synthesized medical data, respectively. 
The table below shows great performance improvements across 3 medical benchmarks with our data.


\begin{table}[t]
\centering
\begin{minipage}{0.44\textwidth}
    \centering
    \caption{\textbf{Type-token ratio statistics}. Type-token ratio (TTR) analysis of LLaVA-NeXT SFT data and \Method-500k.}
    \vspace{-.2em}
    \scalebox{0.95}{
        \begin{tabular}{lcc}
            \toprule
            Metric & LLaVA-NeXT & \Method-500k \\
            \midrule
            Instruction TTR & 0.0018 & 0.0124 \\
            Response TTR & 0.0064 & 0.0086 \\
            \bottomrule
        \end{tabular}
    }
    \label{tab:ttr}
\end{minipage}
\hfill
\begin{minipage}{0.52\textwidth}
    \centering
    \caption{\textbf{\Method~in medical}. Medical data generated by \Method~leads to consistent improvements across 3 medical benchmarks.}
    \scalebox{0.95}{
        \begin{tabular}{lccc}
            \toprule
            Data & PathVQA & VQA-RAD & SLAKE(en) \\
            \midrule
            4k LLaVA & 34.0 & 47.5 & 49.0 \\
            2k LLaVA + 2k Oasis & \textbf{41.2} & \textbf{48.1} & \textbf{56.7} \\
            \bottomrule
        \end{tabular}
    }
    \label{tab:medical}
\end{minipage}
\end{table}

\section{Prompts for Data Filtering}

\Method~contains 2 steps of data filtering: data categorization and quality control. We carefully design the filtering logic to ensure the validity and quality of the data. The efficacy of the filtering process is crucial for the success of our method. Therefore, we handcraft specific prompts for each filtering step to make sure the data is correctly categorized and rated in multiple dimensions. The following are the prompts used in the data filtering process.

\subsection{Data Categorization}

The data generated by the first step of \Method~can be generally categorized into 2 types: image caption and instruction, and only the latter should be retained. Because of the nature of our method, it is observed that the instruction can often hide in large tracts of text, and it is often followed by an answer, which is undesirable. Moreover, the instruction can be in various forms, such as interrogative sentences, imperative sentences, multiple-choice questions, etc. In order to achieve better categorization accuracy, we leverage few-shot examples and design the following prompts.

\begin{lstlisting}
You will be given a text regarding an image. Your task is to determine whether the text contains any instructions. If it contains instructions, extract one instruction. You should extract the instruction, as well as any relevant contextual information that aids in understanding the instruction. 

NOTE:
1. The instruction may take the form of an interrogative sentence, an imperative sentence, a multiple-choice question, or other similar structures. Please identify carefully!
2. Extract ONLY the original instruction, WITHOUT extracting any answers.
3. If the instruction is a multiple-choice question, you should extract the question and the options.
4. If there are multiple instructions, you should extract only one instruction.

You MUST answer with the following format:
Instruction: [an instruction]

If it doesn't contain any instructions, output 'NO_INST'.

----- Example 1:
Text: 
1. Answer the following questions based on the text:\n\n    a. Who increased the number of insurgents in the valley? \n\n    b. When did Singh come to power? What act did he implement?\n\n    c. What is the purpose of the SC-ST Act?

Answer: 
Instruction: Who increased the number of insurgents in the valley?

----- Example 2:
Text: 
There is an animal behind the fence who is holding a bottle.

Answer: 
NO_INST

----- Example 3:
Text: 
In this problem, we have an elephant image that includes several lines and curves.\n\nWe want to transform this image into another animal using the least number of changes.\n\nPlease provide some suggestions on how to achieve this transformation with minimal effort.

Answer: 
Instruction: We want to transform this image into another animal using the least number of changes.\n\nPlease provide some suggestions on how to achieve this transformation with minimal effort.

----- Example 4:
Text: 
Could you please summarize the mission statement of the company and the benefits it promises to its customers in 30 seconds or less?\n The mission of our company is to provide innovative tech solutions for all your needs. We prioritize security and privacy for our users and are committed to excellence. With us by their side, customers can expect a simplified tech journey that feels more defined.

Answer: 
Instruction: Could you please summarize the mission statement of the company and the benefits it promises to its customers in 30 seconds or less?

----- Example 5:
Text: 
I would like to make a real estate agency website using HTML, CSS, and JavaScript.

Answer: 
Instruction: I would like to make a real estate agency website using HTML, CSS, and JavaScript.

----- Example 6:
Text: 
The scene has a window on the top left, a fire hydrant on the bottom right, and two signs in the middle right.

Answer: 
NO_INST
----- End of Example

[Begin of Text]
{text}
[End of Text]
\end{lstlisting}

\subsection{Instruction Quality Control}

The quality control stage evaluates the comprehensive quality of the instructions, including solvability, hallucination, clarity, and nonsense. This step directly determines the representation ability of the data and thus the performance of the model, so the quality control process should be effective and rigorous. We notice that models have a strong tendency to score a 5 with plain prompt, which could lead to biased and insufficient filtering. Therefore, we list the specific scoring criteria for each score in the final prompt.

\paragraph{Prompt for solvability.}~~~

\begin{lstlisting}
Your task is to evaluate the solvability of a query to an image. The solvability can be quantitatively evaluated on a scale of 1 to 5, based on the presence of sufficient information within the image to formulate a complete answer. 

Here are the criteria:

Score 1 (Very Low Solvability): The image contains minimal or no relevant information related to the question, making it nearly impossible to derive a meaningful answer.

Score 2 (Low Solvability): The image provides some information, but key elements are missing, resulting in significant uncertainty.

Score 3 (Moderate Solvability): The image contains a reasonable amount of information that may lead to an answer, but ambiguities or lack of clarity hinder definitive conclusions.

Score 4 (High Solvability): The image offers substantial information that strongly supports answering the question, with only minor uncertainties remaining.

Score 5 (Very High Solvability): The image is rich in detail and clarity, providing all necessary information to answer the question comprehensively.

Please rate the query on a scale of 1 to 5. Use "[[1]]", "[[2]]", "[[3]]", "[[4]]", "[[5]]" to indicate your evaluation score in the key 'Score'.

[Query]
{query}
\end{lstlisting}

\paragraph{Prompt for hallucination.}~~~
\begin{lstlisting}
Your task is to evaluate whether a query to an image contains hallucination content. The determination of whether a question related to an image contains hallucinations can be assessed on a scale of 1 to 5. This scale evaluates the alignment between the question's content and the actual content of the image, identifying discrepancies that indicate hallucinations.

Here are the criteria:

Score 1 (Severe Hallucination): The question bears little to no relation to the image content, filled with substantial errors or completely unrelated information. The discrepancies are so pronounced that they render the question fundamentally flawed in context to the image.

Score 2 (Significant Hallucination): The question diverges considerably from the image, containing multiple erroneous statements or irrelevant details. The inaccuracies are significant enough that they compromise the integrity of the inquiry.

Score 3 (Moderate Hallucination): The question and image content have notable inconsistencies, with several inaccuracies present. While some relevant information is shared, the question includes errors that could lead to misleading conclusions.

Score 4 (Minor Hallucination): The question is largely consistent with the image, but there are minor discrepancies or inaccuracies that do not significantly alter the overall interpretation. These could include slight misinterpretations of color or detail that do not affect the main subject.

Score 5 (No Hallucination): The question aligns perfectly with the image content, containing no errors or irrelevant information. All aspects of the inquiry are directly supported by clear and accurate details within the image.

Please rate the query on a scale of 1 to 5. Use "[[1]]", "[[2]]", "[[3]]", "[[4]]", "[[5]]" to indicate your evaluation score in the key 'Score'.

[Query]
{query}
\end{lstlisting}

\paragraph{Prompt for clarity.}~~~
\begin{lstlisting}
Your task is to evaluate the clarity of a query to an image. The clarity of a question derived from an image can be evaluated on a scale of 1 to 5, reflecting how precisely the question conveys its intent and whether it allows for a definitive answer.

Here are the criteria:

Score 1 (Very Unclear): The question is exceedingly vague and unclear, with multiple interpretations possible. It fails to convey a coherent intent, resulting in uncertainty and an inability to arrive at a definitive answer.

Score 2 (Unclear): The question is largely ambiguous, making it difficult to discern its exact intent. The vagueness significantly hinders the ability to provide a clear answer, leading to potential misinterpretations and disagreements.

Score 3 (Moderately Clear): The question exhibits noticeable vagueness that may cause some confusion. While there are identifiable elements, the lack of precision can lead to varying interpretations and uncertainty in answering.

Score 4 (Clear): The question is generally clear but may contain minor ambiguities that could lead to slight misinterpretations. However, the overall intent remains understandable, allowing for a reasonably definitive answer.

Score 5 (Very Clear): The question is exceptionally clear, leaving no room for ambiguity. It conveys its intent explicitly, and the required answer is straightforward and unambiguous, making it easy to interpret.

Please rate the query on a scale of 1 to 5. Use "[[1]]", "[[2]]", "[[3]]", "[[4]]", "[[5]]" to indicate your evaluation score in the key 'Score'.

[Query]
{query}
\end{lstlisting}

\paragraph{Prompt for nonsense.}~~~
\begin{lstlisting}
Your task is to evaluate whether a query to an image contains nonsense. The presence of nonsense in a question related to an image can be assessed on a scale of 1 to 5.

Here are the criteria:

Score 1 (Severe Nonsense): The question is completely nonsensical, filled with severe grammatical issues, strange characters, or illogical phrases that render it unintelligible. It fails to convey any meaningful intent.

Score 2 (Significant Nonsense): The question is largely incoherent, containing multiple grammatical errors or strange characters that obstruct its meaning. Understanding the question is challenging and may lead to misinterpretations.

Score 3 (Moderate Nonsense): The question exhibits noticeable issues with clarity, such as awkward constructions or vague expressions. While some meaning is still discernible, these factors may lead to confusion.

Score 4 (Minimal Nonsense): The question is generally clear but may contain minor grammatical errors or awkward phrasing that slightly detract from its coherence. These issues do not significantly impede understanding.

Score 5 (No Nonsense): The question is coherent, grammatically correct, and free from any strange characters or phrases. It conveys its intent clearly and logically, allowing for a straightforward understanding.

Please rate the query on a scale of 1 to 5. Use "[[1]]", "[[2]]", "[[3]]", "[[4]]", "[[5]]" to indicate your evaluation score in the key 'Score'.

[Query]
{query}
\end{lstlisting}




\end{document}